%% file: Background.tex
\documentclass[conference]{IEEEtran}
\IEEEoverridecommandlockouts
\usepackage{cite}
\usepackage{amsmath,amssymb,amsfonts}
\usepackage{algorithmic}
\usepackage[linesnumbered,ruled]{algorithm2e}
\usepackage{amsmath}
\usepackage{graphicx}
\usepackage{textcomp}
\usepackage{xcolor}
\usepackage{color}
\usepackage{authblk}
\usepackage{fancyhdr}
\usepackage{booktabs}
\usepackage{bm}
\def\BibTeX{{\rm B\kern-.05em{\sc i\kern-.025em b}\kern-.08em
    T\kern-.1667em\lower.7ex\hbox{E}\kern-.125emX}}
\begin{document}
\title{Costly Features Classification using Monte Carlo Tree Search }
\date{April 2020}

\include{macros}

\maketitle

\begin{abstract}
We consider the problem of costly feature classification, where we sequentially select the subset of features to make a balance between the classification error and the feature cost. In this paper, we first cast the task into a MDP problem and use Advantage Actor Critic algorithm to solve it. In order to further improve the agent's performance and make the policy explainable, we employ the Monte Carlo Tree Search to update the policy iteratively. During the procedure, we also consider its performance on the unbalanced dataset and its sensitivity to the missing value. We evaluate our model on multiple datasets and find it outperforms other methods.
\end{abstract}

\input{intro}
\section{Introduction}
In traditional machine learning tasks, we usually assume that the dataset is complete and all the features are freely available, ignoring the fact that there are certain costs for acquiring all the features. Here, the meaning of cost is generalized beyond financial cost. The computing time, energy consumption, the human resource, and patients' complaints can also be incorporated into this concept. Naturally, in real-world classification tasks, people often want to minimize the cost of feature acquisition, while at the same time achieve good performance. Consider, for example, the task of medical diagnosis. A doctor often approaches the problem by starting with a few symptoms that are easy to check. Based on the result from the initial check on patients, the doctor will further decide whether more advanced examinations are required or not in order to give an accurate diagnosis. When deciding on which examinations to be conducted, the doctor needs to consider not only the cost of each examination but also the relevance to the result. With more information, the doctor can gradually narrow down to set of the potential diseases until he/she is confident to make a conclusion.

 Several scenarios also frequently occur in the domain of computer network optimization, network traffic and energy distribution. These tasks lead to the problem of cost features classification. Generally, our goal is to find an algorithm that can actively acquire useful features to minimize the classification error, while reducing the cost. In particular, different samples could adopt a different subset of features, which is more flexible than the traditional method. Conventional methods can be divided into the following three categories: 
$\newline$
$\newline$
\textbf{Generative Modeling}: In a Bayesian setting, probabilistic models are constructed to evaluate the efficiency of each feature\cite{chen2015value}\cite{chen2014algorithms}. Generally, all these methods need to estimate a probability likelihood that a certain feature will be included based on the features collected previously. Thus, these methods are computationally expensive\cite{chen2014algorithms}, especially in the high-dimensional case. Although it seems efficient, it is hindered by the binary feature assumption.
$\newline$
$\newline$
\textbf{Discriminative Learning Approaches}: Motivated by the success of discriminative learning, these models are usually extended from the traditional machine learning models by considering the cost. Essentially, the predetermined structures limit their performance.  In the TEFE algorithm\cite{liu2008tefe}, a sequence of SVMs is trained for each test object cascadingly. However, it adopts a myopic strategy, ignoring the performance of the entire system. For the tree-based algorithms, Nan proposes Adapt-Gbrt\cite{nan2017adaptive}, which is developed based on random forest (RF). It has a pretrained RF as the High-Performance Classifier (HPC) and adaptively approximates it by training a Low-Prediction Cost model (LPC). Although efficient, the published algorithm can only be used with binary class datasets. Also, the Budget-prune\cite{nan2017adaptive} is proposed to prune the existing RF to make a cost and accuracy trade-off. While this method outperforms many of the previous works, it limits the low-cost classifier to the tree-based model. A recent work focuses on constructing decision trees over all attribute sets, and using the leaves of the tree as the state space of the Markov Decision Process (MDP)\cite{maliah2018mdp}. The value-iteration method is applied to solve the MDP problem. One significant drawback is that it can only be used on small datasets with less than 20 features due to the extremely large number of states in the MDP.
$\newline$
$\newline$
\textbf{Reinforcement Learning}: Intuitively, our task could be modeled as a sequential feature acquisition and classification problem. Starting from the work of\cite{dulac2011datum} , reinforcement learning-based algorithms are proposed. In their work, they directly formalize the problem as a MDP and solve it with linearly approximated Q-learning. Actually, it can only slightly improve the performance  due to the limitation of the linear model. Recently, Janisch replaced the linear approximation with neural network and it performs well on multiple datasets\cite{janisch2019classification}. However, to improve the performance, it requires another pretrained High Performance Classifier, which is impossible to obtain in many real tasks.  Besides, it fails to converge on incomplete or unbalanced datasets because their strategy depends on the Q function, which is sensitive to the missing value and unbalanced data.

In this study, we proposed a DRL based Monte Carlo Tree Search\cite{browne2012survey} method for cost features classification. Unlike DRL-based methods, our method combines the stability of the advantage actor critic(A2C)\cite{babaeizadeh2016reinforcement} algorithm and the efficiency of MCTS in a proper way. In particular, the key contributions of our paper are as follows:

\begin{itemize}
\item \textit{Actor-Critic Pretrain:} The former Q learning based algorithm is problematic. We formulate the cost features classification into MDP and leverage the advantage actor critic algorithm to tackle this problem, where an actor network will guide the agents to select actions based on the stochastic probability, and a critic network will evaluate the actions and guide the actor to a wiser policy.

\item \textit{A Better Reward function:} Besides, considering unbalanced datasets, we also design a specific reward function to deal with it.

\item \textit{Monte Carlo Tree Search:} 
MCTS is employed to further improve the policy learned by A2C. We use a neural network to guide the Monte Carlo tree, and train it iteratively. We also explain the feature acquisition and decision mechanism by MCTS.

\item \textit{Experiment Result: }
We show the superiority of our method on diverse real datasetes compared to other methods.

\end{itemize}

\section{Background}
\subsection{Advantage Actor Critic Algorithm}
The Actor-critic method combines the advantages of both the critic-only methods\cite{chen2019learning} and actor-only methods. On the one hand, the critic's estimation of the temporal-difference(TD) error during the process allows the actor to update with lower variance gradients, which accelerates the learning process. On the other hand, the actor can directly output the target action and parameterize the policies over which optimization procedures can be used directly, endowing the actor-critic algorithm with good convergence properties. 

Following the MDP framework, the goal of the agent is to maximize the expected future reward. Thus, the objective function is defined as follows: $J(\theta)=\sum_{s \in S} d^{\pi_{\theta}}(s) \sum_{a \in A} \pi_{\theta}(s, a) R_{s, a},$ where $S$ and $A$ are state space and action space respectively, $\pi _\theta$ denotes the current policy learnt by the policy network, $ d^{\pi_{\theta}} $ is the stationary distribution of Markov chain under $\pi_{\theta}$ and  $R_{s, a}$ is the reward we get when we do the action $a$ at state $s$. In the advantage actor-critic algorithm, the actor adopts a similar method utilized in the policy gradient algorithm to update the policy network. It moves  $J(\theta)$ toward the local maximum by updating the parameter $\theta$ along the inverse direction of the policy gradient. In order to link the policy gradient with objective function $J(\theta)$, we rewrite $J(\theta) = E_{\pi_{\theta}}[r]$ , where $r=R_(s,a)$ and find the corresponding policy gradient as follows:
\[
\begin{aligned} \nabla_{\theta} J(\theta) =E_{\pi_{\theta}}\left[\nabla_{\theta} \log \pi_{\theta}(s, a) Q^{\pi_{\theta}}(s_t, a)\right] \end{aligned}
\]
Here, $Q$ function means the expected value of the sum of the future rewards when we do the action $a$ at state $s_t$ based on policy $\pi_\theta.$ We extend the one-step MDP to multi-step MDP,  where $ Q^{\pi_{\theta}}(s_t, a)=E[r_t+\gamma \mathop{\max}\limits_{a'} Q^{\pi_{\theta}}(s_{t+1}, a) ].$

Estimating the $Q$ function with the Monte Carlo method may introduce high variance despite its unbiasedness. Consequently,  to reduce the variance of the policy gradient, the advantage actor-critic algorithm uses an evaluation mechanism by a critic network to learn $Q^{\pi_{\theta}}(s_t, a)$. The idea behind is that if the long term value of the state $Q^{\pi_{\theta}}(s_t, a)$ can be approximated appropriately and be adopted to strengthen the policy, the policy can be improved accordingly, which also applies with
the fundamental idea of the actor-critic algorithm. In practice, the critic updates the parameters $\phi$ by estimating the value of the long term value function $Q_{\phi}(s, a) \approx Q^{\pi_{\theta}}(s, a)$. Then, the actor updates the unknown parameters in the policy through feedback from the critic. Now, the policy gradient becomes:

\begin{equation}
\nabla_{\theta} J(\theta)=E_{\pi_{\theta}}\left[\nabla_{\theta} \log \pi_{\theta}(s, a) Q_{\phi}(s, a)\right].
\end{equation}

To further reduce the valiance, we consider replacing the $Q$ function with its advantage over a baseline. It is found that subtracting a baseline function $B(s)$, which only has the relationship with the state instead of action from the policy gradient, does not change the expected value. Besides, the variance will significantly decrease. Naturally, we set $B(s)=V^{\pi_{\theta}}(s)$.  Then, policy gradient will become: $\nabla_{\theta} J(\theta) =E_{\pi_{\theta}}\left[\nabla_{\theta} \log \pi_{\theta}(s, a) A^{\pi_{\theta}}(s, a)\right],$ where $A^{\pi_{\theta}}=Q^{\pi_{\theta}}(s, a)-V^{\pi_{\theta}}_{\phi}(s)$ is the advantage
function. To simplify the calculation, we use the TD error to estimate the advantage.
Finally, the exact policy gradient is obtained as follows:
\begin{equation}
\nabla_{\theta} J(\theta)=E_{\pi_{\theta}}\left[\nabla_{\theta} \log \pi_{\theta}(s, a) \delta^{\pi_{\theta}}\right],
\end{equation}
where $\delta^{\pi_{\theta}}\approx A^{\pi_{\theta}}=r+\gamma V^{\pi_{\theta}}_{\phi}(s_{t+1})-V^{\pi_{\theta}}_{\phi}(s_t).$

\section{Problem Formulation and MDP Framework}
In this section, we introduce the problem formulation that translates the task of cost features classification as Markov Decision Process (MDP). Traditionally, MDP models decision making process in a discrete-time stochastic environment. For the problem of sensitive classification, by considering the feature selection and label prediction as a sequential decision making process, we can model the problem  under the framework of MDP then solve it \textit{via} reinforcement learning. Specifically, given a data sample needed to be classified, by selecting different features to classify, the process is akin to an agent interacting with an environment to achieve certain goals. The goal in our task is to achieve accurate classification with minimal number of features. By training such an agent using labelled data, we can make a prediction with the learned policy.

Consider a data sample $(\mathbf{x}, y)$, where $\mathbf{x} \subset \mathbb{R}^p$ is a feature vector $\mathbf{x}=(x_1,x_2,...,x_p)$ and $y$ is the corresponding class label. To predict $y$, we need to inspect (a.k.a., acquire) a set of features to make the decision. We denote the cost of acquiring a feature $x_p$ as $c_p \in \{c_1,...,c_p\}$. Under the framework of MDP, the agent needs to take a sequence of actions to interact (i.e., acquire or not) with the data features in order to make the final classification decision. By beginning with zero features, at each step, the agent takes an action to either acquire additional feature or make a label prediction and terminate the process (i.e., end of episode). For an arbitrary step $t$,  we maintain a feature set $\mathcal{A}_t$ which includes the features that are acquired and also record the corresponding cost $C_t$ for acquiring these features ($\mathcal{A}_0\in\emptyset, C_0=0$). At the next step $t+1$, if the agent decides to acquire another feature $x_i|i \in \{1, \ldots, p\}$, then the variables get updated as $\mathcal{A}_{t+1}=\mathcal{A}_t \cup x_i$ and $C_{t+1} = C_t + c_i$. Otherwise, the agent will make its final prediction on the class label and terminate. Overall, the agent needs to learn a policy that minimizes the prediction error as well as the feature cost. We can formulate a loss function as shown below:  

\begin{align}
\mbox{argmin}\frac{1}{|A_T|}\sum\limits_{(x,y)\in A_T} L(f_{\theta}(x,A_t),y)+\lambda C_T
\end{align}

\noindent where $L$ denotes the classification error while $\lambda$ serves as a scaled factor balancing the importance of classification error and the total cost. Notably, although this problem presents as an optimization problem, it is analytically intractable because deciding the best feature set is usually NP-hard. However, by formulating the problem under MDP framwork, we can solve it by using reinforcement learning (dynamic programming). Specifically, given the MPD 4-tuple $(\mathbf{S},\mathbf{A}, \mathbf{R},\mathbf{P})$, we layout the definition of the 4 components (i.e., state, action, reward and transition) as follows:

\textbf{$\mathbf{S}$ - State:}
In our task, the state should summarize the information of the current feature subset at each step. Hence not only the index of the feature is important, but also the value. We borrow the idea from (cit) and construct the t-step state vector $s_t$ as a concatenation of two vectors $I_t$ and $x_t$. The vector $x_t$ is given as $x_t[i]=x[i]$ if $f_i\in A_t$; otherwise $x_t[i]=0$. The indicator encoding $I_t$ illustrates the index of the selected features. The indicator vector is necessary because it can help the agent to distinguish the zero value of a feature to the indicator of the position. 

\textbf{$\mathbf{A}$ - Action:}
The agent has two choices in selecting one step. Naturally, they can adopt feature acquisition, which means they can still select the unexplored and available features. Besides, one agent can terminate the episode by making a "stop" action, and make a classification based on the final feature subset $A_T$. Furthermore, we didn't use any imputation method to deal with the missing value and only allow the agent to select available features at a time.

\textbf{$\mathbf{R}$ - Reward:}
When the agent collects feature $f_j$ at step $t$, a negative $-\lambda c(f_j)$ will be returned as an instant reward right after the action. Unlike the "feature acquisition" actions, the "terminate" action will lead to a reward reflecting the classification result for this sample. In practice, unbalanced data classification occurs frequently. In order to tackle this problem, the agent, at the terminal, is assigned various misclassification costs to different classes according to a special loss function, in which the misclassification cost of the minority class is larger than that of the majority class.   Thus, we define the reward function for termination as follows:
$$ \mbox{reward}=\left\{
\begin{aligned}
&1\quad \mbox{if}\ f(x,A_t)=l\ \mbox{and}\ i\in D_\mbox{min}  \\
&-1 \quad \mbox{if}\ f(x,A_t)\not=l\ \mbox{and}\ i\in D_\mbox{min}  \\
&\delta \quad \mbox{if}\ f(x, A_t)=l\ \mbox{and}\ i\in D_\mbox{maj}\\
& -\delta \quad \mbox{if}\ f(x, A_t)\not=l\ \mbox{and}\ i\in D_\mbox{maj}\\
\end{aligned}
\right.
$$
where $\delta$ is bounded by [0,1]. In fact, $\delta$ is a trade off parameter to adjust the importance of the majority class. Our model achieves the best performance in experiment when $\delta$ is around the imbalance ratio $\delta=\frac{\#D_{min}}{\#D_{max}}$. Thus, we set the trade off parameter to be the imbalance ratio.

\textbf{$\mathbf{T}$ -Transition:}
In this MDP, when an action a is taken for a particular state $s$, the next state $s'$ is determined. That is, this MDP transition is deterministic.

\section{Training an agent with A2C}
\subsection{Learning and Inference}
In this section, we provide the details of training the agent with A2C algorithm. We basically follow the algorithm described in the background and adopt one core mechanism: policy gradient entropy. Specifically, during the training procedure, the agent generates the episode ($s_0$,$a_0$,$r_1$,$s_1$,$a_1$,$\cdots$,$s_{T+1}$). Here, the immediate reward $r_t$ is obtained from the environment. On one hand, for the 'feature acquisition' action, the agent gets an penalty $\lambda c(f_j)$ after it selects feature $j$. One the other hand, when the action is terminal, the reward will be calculated based on the classification result. 

After generating the trajectories for each sample, the agent is trained in an on-policy manner. All the samples are used to calculate the policy gradient and update the policy. However, the samples might be noisy especially in the beginning. Hence, we introduce the policy entropy $H(\bm{\pi(s)})=E_{a\sim\pi}(-\log\bm{\pi(s)})$ as a regularizer to control the exploration. We minimize the summation of expected reward and the policy entropy when training the network. For the cost classification problem, it is effective to overcome the premature convergence.  
\subsection{Neural Network architecture}
Intuitively, the actor-network and the critic-network should share some information, since both of them make a prediction based on the same state and optimize the objective function. According to the experiment of AlphaGo Zero\cite{silver2017mastering}, a partially sharing model performs better than two separate models. Thus, we adopt the partially shared neural network with three common layers and two branches. During the training procedure, all the layers and branches are jointly trained. The structure of the neural network is shown in figure 1.

\begin{figure}[h!]
\vskip 0.2in
\vspace{-0.5cm}
\setlength{\abovecaptionskip}{-0.3cm}
\setlength {\belowcaptionskip} {-0.1cm}
\begin{center}
\centerline{\includegraphics[scale=0.1]{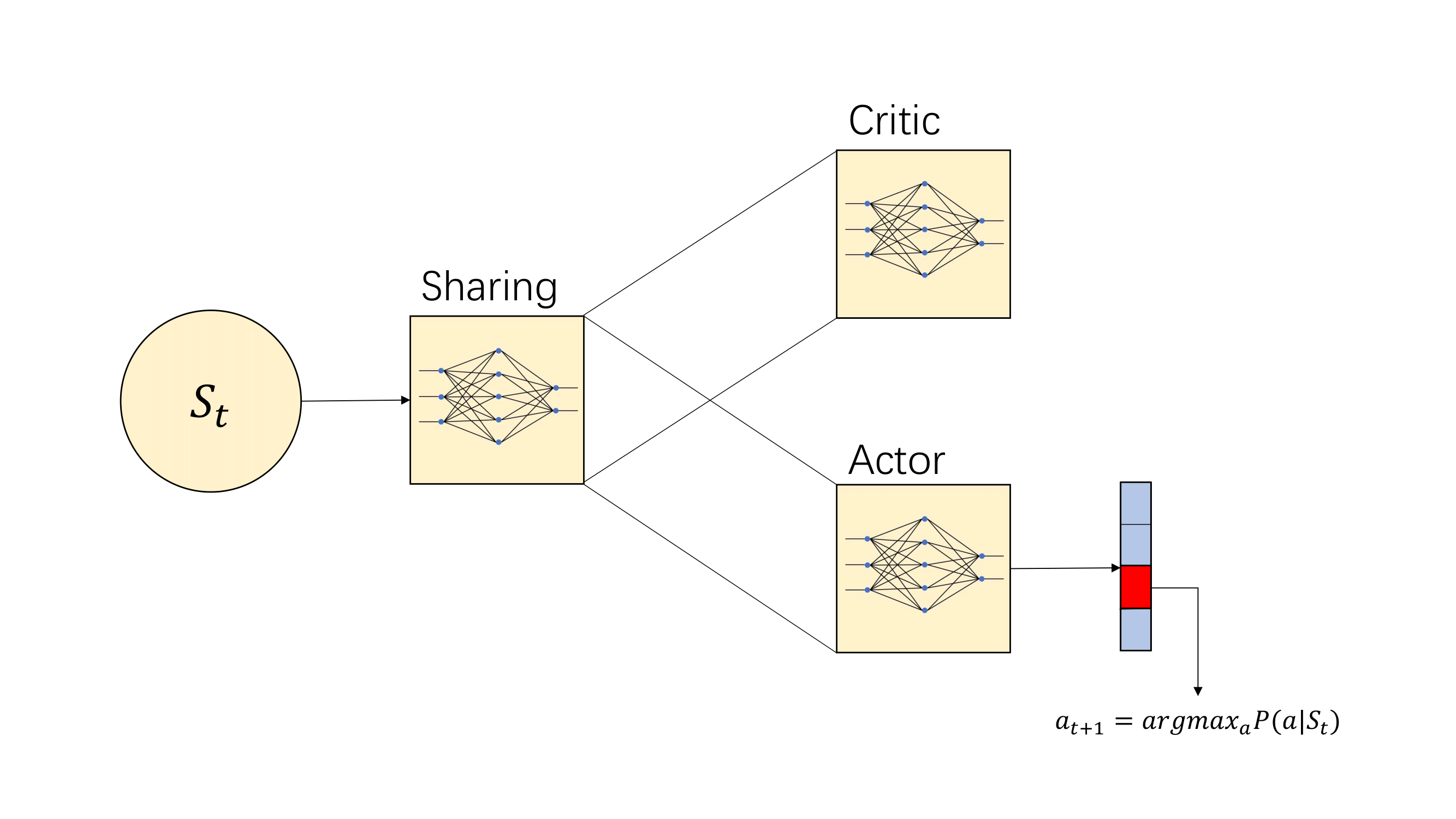}}
\caption{The partially shared neural network} 
\label{fig:diag}
\end{center}
\vskip -0.1in
\end{figure}
\subsection{Training Procedure}
Algorithm 1 is the psuedo code of Advantage Actor-Critic RL Algorithm.

\begin{algorithm}
    \SetKwInOut{Initialized}{Initialized}
    \text{Initialized parameters $\theta$,s,$\phi$}\;
    \Initialized{Actor Network  $\pi_{\theta}(s)$ and Critic Network  $V^{\pi_{\theta}}_{\phi}(s)$}
    Initialized actors and critics learning rate $\gamma_{a}$ and $\gamma_{c}$ and TD error discount factor $\beta$\;
    \For{each training epoch $n=1, 2,..., N$}
      {
        Receive initial state $s_{1}$\;
        \For{each episode $t=1, 2,..., T$}
      {
        Select and Execute $a_t$ based on current state and policy\;
        Received $r_t$ and next state $s_{t+1}$ based on $(a_t, s_t)$\;
        Calculate TD error in critic  $\delta^{\pi_{\theta}}=r+\gamma V^{\pi_{\theta}}_{\phi}\left(s_{t+1}\right)-V^{\pi_{\theta}}_{\phi}\left(s_{t}\right)$\;
        Update $\Delta\phi$ for the critic network\;
        $\Delta\phi$=$\Delta\phi$+$\delta^{\pi_{\theta}}$\;
        Calculate policy gradient in actor:\\ $\nabla_{\theta} J(\theta)=E_{\bm{\pi_{\theta}}}\left[\nabla_{\theta}  \bm{\pi}_{\theta}\left(s_{t}\right) \delta^{\pi_{\theta}}\right]-\nabla_{\theta}H(\bm{\pi}_{\theta}(s_t))$\;
        Update $\Delta\theta$ for the policy network
        $\Delta\theta=\Delta\theta+\nabla_{\theta} J(\theta)$\;
        Update state: $s_{t}=s_{t+1}$\;
        
      }
      Update critic network by $\Delta\phi$\;
      $\phi=\phi+\alpha\Delta\phi$\;
      Update value network by $\Delta\theta$\;
      $\theta=\theta+\beta\Delta\theta$
      \Endfor
      }
      \Endfor
      {
        return\;
      }
    \caption{Advantage Actor-Critic RL Algorithm}
\end{algorithm}

\section{Improving by Monte Carlo Tree Search}
\subsection{Motivation}
Even if the A2C converges, it fails to detect some minor distinctions between features and tends to incorporate irrelevant features. Besides, the performance is unstable. This problem is triggered due to the model-free nature of the advantage actor algorithm. Without planning and the prior knowledge of the environment, the agent is likely to gather some noisy samples, which may lead to over-fitting. Particularly, the agent seems to select irrelevant features for some samples although it performs well on others.

To further fine screen the features, we employ the  Monte Carlo Tree Search (MCTS) to improve the policy learned by A2C. MCTS is a model-based algorithm for decision-making problems notably applied in games such as AlphaGo. In our task, an experiment made on the Wine data set shows that the MCTS could reduce the cost by 0.16 on training dataset compared with the policy network learned by A2C, while maintaining the same accuracy. Thus, MCTS estimates the potential reward of introducing a feature by interaction with the environment and then use it to adjust the policy in the best first nature, allowing us to focus on the most economic feature set.

 Another advantage brought by MCTS is that it illustrates the decision rule in a tree-structure model, by which an explainable decision system can be built.
In areas like healthcare and psychology, explainability is an essential prerequisite of using a model or a system. An overall diagram of the learning algorithm is ploted in Figure 2.
\begin{figure}[h!]
\vskip 0.2in
\vspace{-0.5cm}
\setlength{\abovecaptionskip}{-0.3cm}
\setlength {\belowcaptionskip} {-0.1cm}
\begin{center}
\centerline{\includegraphics[scale=0.3]{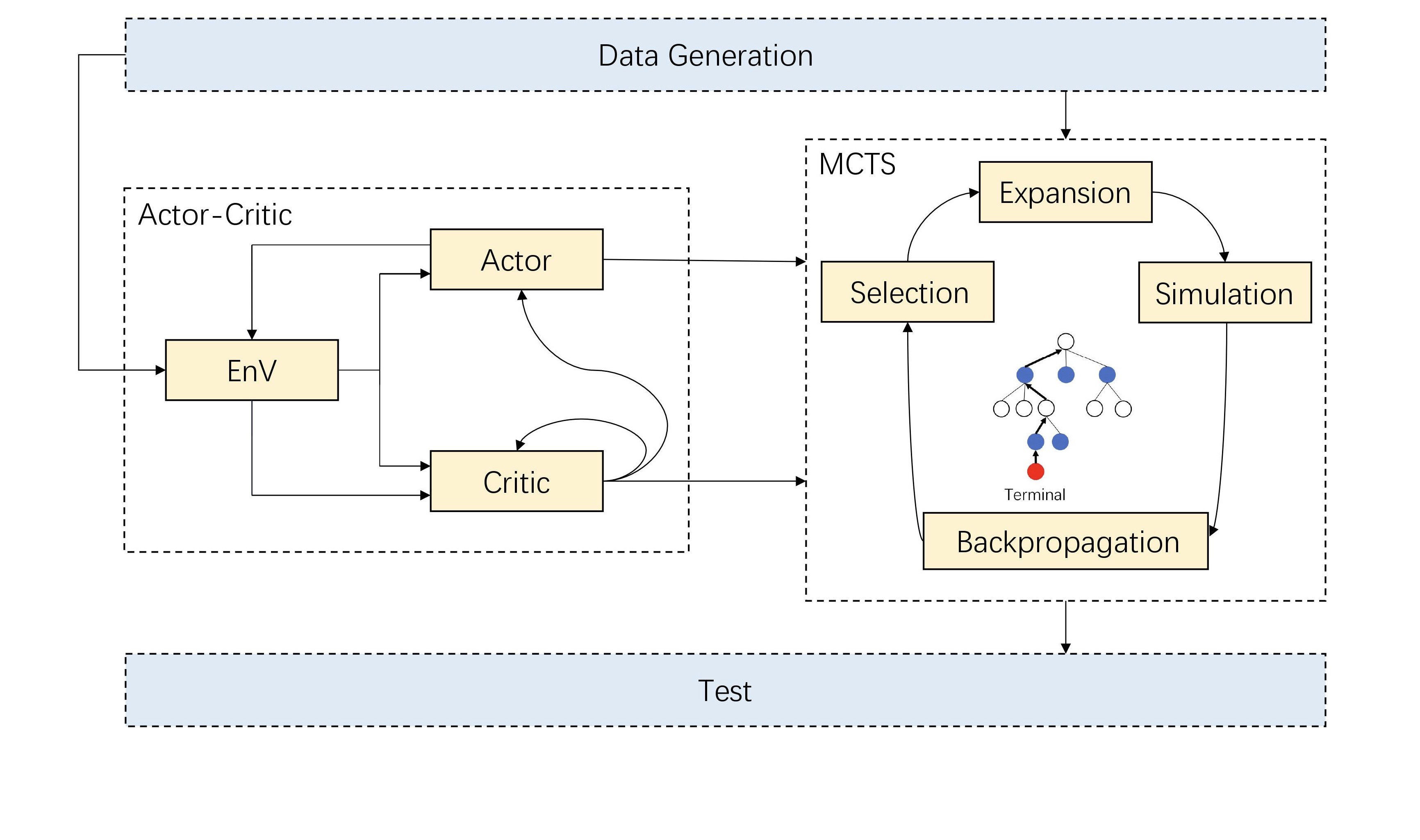}}
\caption{The system diagarm of RL and MCTS} 
\label{fig:diag}
\end{center}
\vskip -0.1in
\end{figure}
\subsection{Neural Monte Carlo Tree Training Procedure}
In this section, we show how the Neural Monte Carlo Tree algorithm can be adapted and applied to cost features classification. The core idea is to iteratively train the neural network $f_{\theta}$. The high-level training algorithm is shown in Algorithm 2.

Similar to the AlphaGo Zero algorithm, the MCTS here is applied as a policy improvement operator to iteratively improve the policy network\cite{huang2019coloring} in training phase. We initialize $f_{\theta}$ with the default neural network obtained from A2C to map the representation of the state $s_t$ to next action probability and a value,
$(\textbf{p},v(s)=f_{\theta}(s))$.
The vector $\textbf{p}$ represents the probability of features and terminal actions to be incorporated at state $s_t$. Here, $v(s_t)$ denotes the potential reward of each action. For each sample $(\textbf{x},y)$ randomly selected from training data, MCTS is adopted to search for a better policy $\bm{\pi}^{M}$ and the corresponding value estimation. These improved policies as well as the values are used as labels to train the neural network, which aims to improve the estimates of the state-value and policy functions, which, in turn, are used for MCTS in the subsequent episodes until it converges.

Specifically, for a given sample $(\textbf{x},y)$, we construct a MCTS according to the step given in Algorithm 2. Each node in the search tree contains two quantities $(Q(s_t, a), N(s_t, a))$: the action value function $Q(s_t,a)$ and visiting count $N(s_t,a)$. Here $Q(s_t,a)$ denotes the expected long-term reward of selecting feature $a$ and $N(s_t,a)$ denotes the number of times feature a is selected at state $s_t$ during the simulation. 
One thing needs to be mentioned is that the neural network produces $v(s_t)$, but the MCTS requires $Q(s_t,a)$. Hence, we use the following method to approximate it.
   \begin{align}
    Q(s_t,a)\approx R(s_t,a)+\gamma V(s_{t+1})
   \end{align}

When the algorithm terminates, we can calculate and collect the value and policy labels for each state to train the neural network. The reward label is already calculated before. For the policy label, we use the empirical frequency of MCTS: $\bm{\pi^{M}}(s)=\frac{N(s,.)}{\sum_{a}N(s,a)}$ to estimate it.

After gathering the improved labels, we train the neural network $f_{\theta}$ to update the policy by minimizing the loss function:

\begin{align}\label{eq:eps}
    l=(r+v(s_{t+1})-v(s_t))^2-\bm{\pi^{M}}\mbox{log}(\bm{p})
\end{align}
where a) the first term measures the mean-square-error MSE of the predictive value function, b) the second term is the KL divergence between the previous policy and the policy improved by MCTS. 
\begin{algorithm}%[H]
\caption{Constructing an MCTS tree for a \textbf{single} $(\bm{x}, y)$ pair} \label{constructing_MCTStree}
\begin{algorithmic}[1]
\STATE \textbf{Initialize}: Initializing the neural network in MCTS with the default neural network from A2C. Set the root state as $r$.
\STATE Run $N$ number of MCTS simulations from $r$ according to PUCT to grow the tree.  
\STATE Select the action/column $a$ with the largest $N(c,a)$. After taking action $a$, the root state $r$ is transferred to a new state $r^{'}$
\STATE Update  $r \leftarrow r'$. 
\IF {$r$ is a terminal state}
\STATE \textbf{return}: all states $r$ visited
\ELSE
    \STATE \textbf{return} to step 2
\ENDIF
\end{algorithmic}
\end{algorithm}

\subsection{Neural Monte Carlo Tree Search}

Details of how the MCTS works in our task requires clarification. Each node in the tree is associated with 4 quantities (s,$Q(s_t,a)$,$\pi(s_t,a)$,$N(s,a)$).  As illustrated in Figure 3, at each iteration, MCTS performs four steps: selection, expansion, simulation, and backpropagation. In particular, starting from the current root state $s_t$, we select action $a_t$ according to  PUCT (Polynomial Upper Confidence bound for Trees):
\begin{figure}[h!]
\vskip 0.2in
\vspace{-0.5cm}
\setlength{\abovecaptionskip}{-0.3cm}
\setlength {\belowcaptionskip} {-0.1cm}
\begin{center}
\centerline{\includegraphics[scale=0.12]{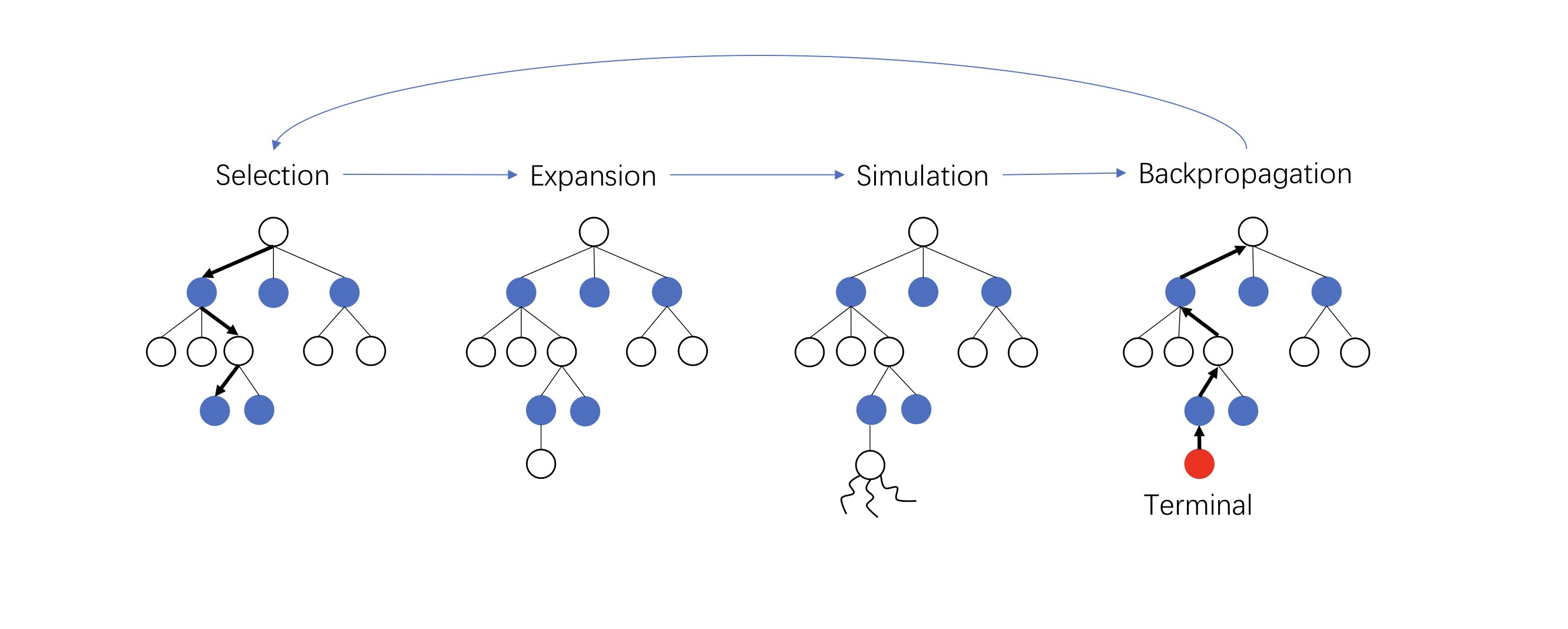}}
\caption{The growth of MCTS} 
\label{fig:diag}
\end{center}
\vskip -0.2in
\end{figure}
\[
a_t = \arg\max_a \left\{ 
        Q(s_t,a)+c \cdot \pi_\theta(s_t,a) \frac{\sqrt{\sum_b N(s_t,b)}}{N(s_t,a) + 1}\right\},
\]
Where $c$ denotes a temperature parameter balancing the exploitation and exploration. Instead of totally relying on the roll-out step, using a neural network to guide the search reduces both the breadth and depth of the space, leading to a significant speedup. Once the agent selects a terminal action or the size of the feature subset exceeds $T$, we stop searching and propagate the obtained reward along the search path back to the root for all the state visited previously. The $Q$ value and the value of $N$ will be updated according to the formula:
\begin{align}
N(s_t, a_t) \leftarrow N(s_t, a_t) + 1 \\
Q(s_t, a_t) \leftarrow \frac{(N(s_t, a_t) - 1) Q(s_t, a_t) + v(s_v)}{N(s_t, a_t)} 
\end{align}

$\newline$
\section{Experiment Details}
\subsection{Experiment on Public Dataset}
\subsubsection{Implementation Details}
To validate the performance of our model, we use several datasets from UCI. All of them are multi-class datasets with number of features ranging from 6-50 and class ranging from 2-10. All the datasets are assigned random costs from 0.1 to 1. We normalize the datsets and randomly split the dataset into training, validation and testing sets. In our experiment, we randomly split the dataset 20 times for one setting and run our method. The final result is obtained by averaging the 20 experiments.
Q denotes the Q learning algorithm proposed by Janish, and similarly to our method, it also apply the reinforcement learning method. In their algorithm,
HPC is used for improving the accuracy by using all features. However, in real problem, HPC is hard to obtained because of the missing value in the dataset, especially in healthcare area. Hence, considering the practicability，we didn't use the HPC for all the methods. In TEFE, multiple stages of
margin based reject classifiers have been considered. It is widely used in the graph classification task. Here we also use it as one of the baseline. All algorithms contain a similar parameter $\lambda$ to balance the accuracy and the cost. We set the $\lambda$ with four setting $\{1,0.1,0.01,0.001\}$ separately. By altering the value of $\lambda$, we can find that higher $\lambda$ forces the agent to prefer lower cost feature and the feature subset is small and vice cersa. Besides, for the Neural Network used in A2C+MCTS and Q, we use the three-layer multiple linear perceptron(MLP) with 256 neurons in each layer.  More result can be found in Table1. 
\subsubsection{Comparison of the performance on normal dataset}

T{\color{red} TableDeleted} shows the accuracy of three algorithms on different settings. Comparing the result, we can found that our method(A2C+MCTS) outperforms the Q learning algorithm on datasets Wine, yeast, Aba, Bld and Cre. Among these results, it seems that the TEFE has the worst performance due to its predetermined structure. The difference between A2C+MCTS and Q is pretty on dataset Aba and Cre, in which 5 percent samples of total contain missing value. Thus, it indicates that our method is more robust to the missing value than Q. 

\subsubsection{Comparison of the performance on unbalanced dataset}
Table2 shows the AUC of three algorithms on the unbalanced datasets. To make the class unbalanced, we modify the proportion of the label by randomly delete the samples with label 1. After modification, the proportion of samples with label 1 equals $\{0.2,0.15,0.1\}$ and we set the $lambda$ equas 0.001.  As shown in the Table2, our specific designed reward is beneficial for the unbalanced datasets.

\subsubsection{Show the decision rule on public dataset}
As the policy is finally improved by the MCTS, we can directly extract the decision rule from the MCTS by selecting the most frequently visited path during the training procedure. Although the MCTS efficiently construct the whole tree and focus on the important paths, the complete tree still contains a lot of irrelevant path. Hence, here we only select the path leading to the leaf with at least 50 times visits.
We focus on the decision rule on dataset Spe. As illustrated in figure 4, we can find the most frequent strategy the actor network use. In the beginning, the agent select variable1 and then acquires three different variables according to the value of variable1. Different from the binary structure of the decision tree, the number of branches from one node is not fixed, which has a far more flexible structure than decision tree. Furthermore, the decision rule set by MCTS can help people explore the importance and 
efficiency of different features.

{\color{red}aaa}

\begin{figure}[h!]
\vskip 0.2in
\vspace{-0.5cm}
\setlength{\abovecaptionskip}{0.1cm}
\setlength {\belowcaptionskip} {-0.5cm}
\begin{center}
\centerline{\includegraphics[scale=0.5]{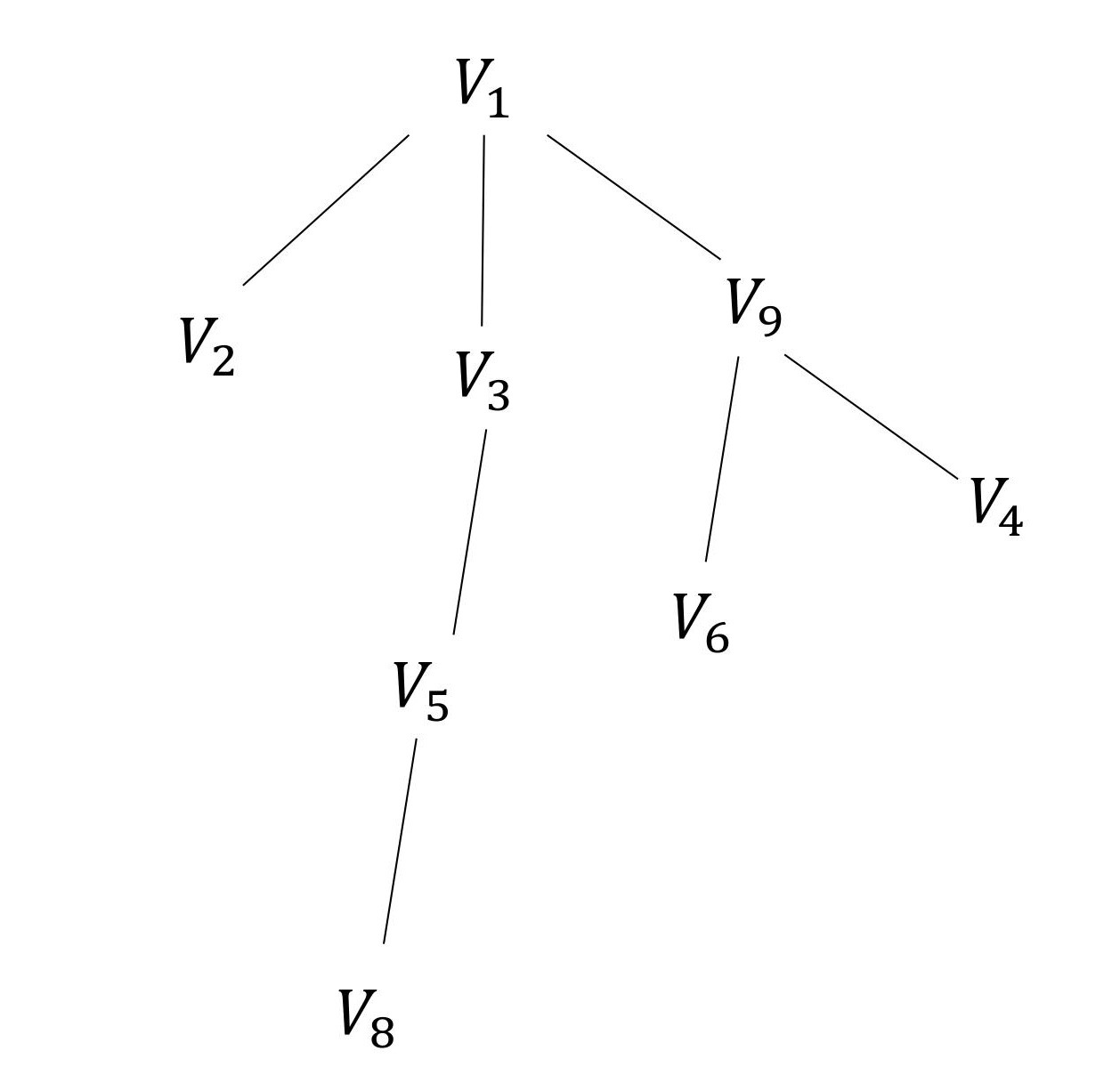}}
\caption{Decision Rule of SPE} 
\label{fig:diag}
\end{center}
\vskip -0.2in
\end{figure}

$\newline$

\begin{tabular}{ccccc}
\toprule  
Dataset& Number of& Class& Number of& Source\\
& Features& & observations& \\
\midrule  
Wine& 13& 3& 70& UCI\\
Yeast& 8& 10& 600& UCI\\
Aba& 7& 2& 4177& UCI\\
Bld& 6& 2& 345& UCI\\
Cre& 10& 3& 1000& UCI\\
Spe& 44& 2& 267& UCI\\
\bottomrule
\caption{Table 1}
\end{tabular}

\vspace*{2\baselineskip} 

\begin{tabular}{cccc}
\toprule  
& A2C+MCTS& Q& TEFE\\
\midrule  
0.2& 0.73& 0.67& 0.58\\
0.15& 0.58& 0.55& 0.39\\
0.1& 0.53& 0.49& 0.31\\
\bottomrule
\caption{Table 2}
\end{tabular}

\section{Conclusion}
We propose an A2C+MCTS two stage algorithm for the cost features classification problem. Different from other algorithm, we leverage the actor critic framework to improve the stability of the model. To be efficient, we use the MCTS to further improve the policy learnt by A2C. Besides, we also consider the explainability of the model. Further research could be left to explore more efficient algorithm to extract the decision rule from the monte carlo tree.

\bibliographystyle{ieeetr}
\bibliography{ci}

\end{document}

%% file: macros.tex
% Revision with replaced text
\newcommand{\rev}[2]{\sout{#1}\textcolor{red}{#2}}
% Colored revision without replaced text
% \newcommand{\rev}[2]{\textcolor{blue}{#2}}
% Accept revision
% \newcommand{\rev}[2]{{#2}}

\newcommand{\newsub}[1]{\vspace{0.025in}\noindent\textbf{{#1}}}